\documentclass[11pt,a4paper]{article}

\usepackage[utf8]{inputenc}
\usepackage[T1]{fontenc}
\usepackage{amsmath,amssymb,amsthm,mathtools}
\usepackage{booktabs}
\usepackage{graphicx}
\usepackage{hyperref}
\usepackage{cleveref}
\usepackage{xcolor}
\usepackage{microtype}
\usepackage[margin=1in]{geometry}
\usepackage{natbib}
\usepackage{algorithm}
\usepackage{algpseudocode}
\usepackage{caption}
\usepackage{subcaption}
\usepackage{bm}
\usepackage{enumitem}

\DeclareMathOperator{\diag}{diag}

\DeclareMathOperator{\E}{\mathbb{E}}

\newcommand{\vb}{\mathbf{v}_{\mathrm{b}}}
\newcommand{\vswitch}{\mathbf{v}_{\mathrm{sw}}}
\newcommand{\R}{\mathbb{R}}
\newcommand{\inner}[2]{\langle #1,\, #2 \rangle}
\newcommand{\norm}[1]{\lVert #1 \rVert}
\newcommand{\abs}[1]{\lvert #1 \rvert}

\newtheorem{definition}{Definition}

\newtheorem{remark}{Remark}

\title{Optimizer-Induced Low-Dimensional Drift and\\Transverse Dynamics in Transformer Training}
\author{Yongzhong Xu\thanks{\texttt{abbyxu@gmail.com}.
  \quad Code: \url{https://github.com/skydancerosel/mini_gpt}}}
\date{}

\begin{document}
\maketitle

\begin{abstract}
We analyze cumulative parameter trajectories of transformer training under
AdamW and identify a dominant low-dimensional drift direction (``backbone'')
that captures 60--80\% of long-horizon displacement from initialization.  This
direction is highly stable over rolling training windows yet reorients gradually
across phases, particularly following objective reweighting.  Per-batch
gradients exhibit near-noise-floor alignment with the backbone, whereas
optimizer-integrated updates align strongly with it, indicating that the
structure emerges from accumulated optimizer dynamics rather than instantaneous
gradient geometry.

Replacing AdamW with SGD-family optimizers eliminates this structure, and
reducing $\beta_2$ smoothly degrades backbone dominance and reheating
recoverability.  Reheating experiments show that transverse probe modes can be
transiently re-excited without substantially altering accumulated backbone
drift.

These results provide a trajectory-level characterization of optimizer-induced
geometric structure in transformer training and shift attention from
instantaneous gradient properties to cumulative update dynamics.
\end{abstract}

\section{Introduction}\label{sec:intro}

Training dynamics in deep neural networks are typically analyzed through the
geometry of the loss landscape: curvature, sharpness, and stochastic gradient
noise are taken to determine how optimization proceeds.  While this perspective
captures important local properties of learning, it emphasizes instantaneous
gradient structure rather than the accumulated trajectory of parameters over
long training horizons.

In high-dimensional models, these two viewpoints need not coincide.  Per-step
gradients may be large and highly variable, yet their cumulative displacement
can concentrate in a small number of coherent directions.  Understanding this
cumulative geometry is essential for characterizing optimizer-induced implicit
bias and long-horizon training behavior.

In this work, we study the global parameter trajectory of transformer training
under AdamW~\citep{loshchilov2019decoupled}.  Rather than analyzing local
curvature or single-step gradients, we examine cumulative displacement from
initialization across checkpoints.  We find that training admits a dominant
low-dimensional drift direction---which we term the \emph{backbone}---that
captures the majority of cumulative parameter movement.  Across blocks and
seeds, the first principal component of uncentered trajectory PCA explains
60--80\% of total drift.  This direction is highly stable over rolling training
windows yet reorients gradually across longer horizons, particularly following
objective reweighting.

Crucially, the backbone is not aligned with instantaneous gradient directions
and is nearly orthogonal to leading Fisher curvature
modes~\citep{martens2020new}.  Per-batch gradients are close to isotropic
relative to the backbone.  However, the optimizer-integrated update---after
momentum accumulation and adaptive per-parameter
normalization~\citep{kingma2015adam}---exhibits strong alignment with it.  This
indicates that the backbone is not a property of the loss landscape alone,
but emerges from accumulated optimizer dynamics.

To test this interpretation, we replace AdamW with SGD-family optimizers while
holding model, data, and schedule fixed.  Under SGD with or without momentum,
trajectories remain nearly colinear and fail to develop the multi-dimensional
structure observed under AdamW---even at matched validation loss.  This
establishes that the backbone is optimizer-induced rather than a generic feature
of the objective.

We further examine the dynamical consequences of this structure.  Oscillatory
regime switching between competing objectives occurs primarily in directions
transverse to the backbone.  Reheating experiments show that these transverse
modes can be transiently re-excited from late-training checkpoints without
substantially altering accumulated backbone drift.  This behavior is consistent
with a slow--fast decomposition: a low-dimensional, optimizer-shaped drift
manifold governs long-horizon evolution, while high-dimensional transverse
dynamics mediate switching.

Together, these findings shift attention from instantaneous gradient geometry to
cumulative trajectory structure.  They provide a concrete empirical
characterization of optimizer-induced implicit bias in transformer training and
suggest that adaptive optimization reshapes not only convergence rates but the
geometry of learning itself.

\paragraph{Relation to prior work.}
The separation of dynamics into slow and fast components has classical roots in
dynamical systems theory, particularly in slow manifold and time-scale
separation results such as Fenichel-type
theorems~\citep{saxe2014exact}.  In optimization and deep learning, related
ideas appear in analyses of momentum methods and adaptive optimizers, where
Adam-type algorithms are understood as inducing effective geometry changes
through preconditioning and sign-consistent
updates~\citep{kingma2015adam,loshchilov2019decoupled,cohen2021gradient}.
Recent work has also emphasized implicit bias and trajectory-level properties of
high-dimensional training
dynamics~\citep{lewkowycz2020large,power2022grokking,frankle2020linear}.

Our contribution differs in emphasis and object of study.  Rather than analyzing
local curvature, stationary points, or instantaneous update rules, we examine
the cumulative geometry of training trajectories and identify a stable drift
direction that dominates long-horizon parameter displacement.  We show that this
backbone direction is not aligned with per-batch gradients or with top curvature
modes, but instead emerges from optimizer-integrated temporal coherence.  This
shifts attention from static loss-landscape structure to trajectory-level
geometry, providing a concrete empirical characterization of optimizer-induced
slow-manifold behavior in transformer training.

\section{Experimental Setup}\label{sec:setup}

\subsection{Model and Data}

We train a decoder-only Transformer~\citep{vaswani2017attention} in the GPT-2
family~\citep{radford2019language}: 8 layers, $d_{\mathrm{model}}=512$, 16
attention heads, $d_{\mathrm{ff}}=2048$, totalling 51M parameters.  The
training corpus is TinyStories~\citep{eldan2023tinystories}.  With probability
$p_{\mathrm{probe}}=0.10$, a training sequence is replaced by a probe sequence
containing a codeword--value pair; the model must predict the value token given
the codeword at out-of-distribution gap distances.

\subsection{Training Configuration}

\begin{table}[h]
\centering
\caption{Training hyperparameters.}
\label{tab:hyperparams}
\begin{tabular}{@{}ll@{}}
\toprule
\textbf{Parameter} & \textbf{Value} \\
\midrule
Optimizer              & AdamW ($\beta_1{=}0.9$, $\beta_2{=}0.95$) \\
Learning rate          & $10^{-3}$, cosine decay, 1500-step warmup \\
Weight decay           & 0.5 \\
Probe loss weight $\lambda$ & 2.0 (steps 1--3999), 4.0 (steps 4000--10000) \\
Effective batch size   & 128 ($64 \times 2$ gradient accumulation) \\
Total steps            & 10{,}000 \\
Checkpoint interval    & 200 steps (51 checkpoints) \\
Seeds                  & 42, 271 \\
\bottomrule
\end{tabular}
\end{table}

The composite loss at each training step is
\begin{equation}\label{eq:composite-loss}
    \mathcal{L}(\bm{\theta}) \;=\; \mathcal{L}_{\mathrm{LM}}(\bm{\theta})
    \;+\; \lambda\,\mathcal{L}_{\mathrm{probe}}(\bm{\theta}),
\end{equation}
where $\mathcal{L}_{\mathrm{LM}}$ is the standard next-token prediction
cross-entropy and $\mathcal{L}_{\mathrm{probe}}$ is the cross-entropy on the
codeword retrieval task.  The weight $\lambda$ is doubled at step 4000 to
intensify probe competition.

\subsection{Oscillation Phenomenology (Brief)}

Over 10{,}000 steps, the out-of-distribution probe accuracy $p_{\mathrm{ood}}$
oscillates between 0.40 and 0.78 (seed~42) or 0.20 and 0.67 (seed~271), while
the LM validation loss decreases monotonically from ${\sim}10$ to ${\sim}1.2$.
The oscillations damp after step ${\sim}7000$, and the model settles into an
LM-dominant regime ($p_{\mathrm{ood}} < 0.20$).  We set aside the full
oscillation phenomenology and focus on the geometric structure of the underlying
training trajectory.

\section{The Backbone}\label{sec:backbone}

\subsection{Trajectory PCA: One Direction Dominates}\label{sec:pca}

We analyze the cumulative parameter drift using \emph{uncentered} principal
component analysis.  Let $\bm{\theta}(t) \in \R^D$ denote the vectorized
parameters of a single transformer block at checkpoint~$t$, with block
dimensionality $D \approx 3.1 \times 10^6$.

\begin{definition}[Drift matrix]
    The drift matrix $\mathbf{X} \in \R^{T \times D}$ has rows
    \begin{equation}\label{eq:drift}
        \mathbf{x}(t) \;=\; \bm{\theta}(t) - \bm{\theta}(0),
        \qquad t = 1,\ldots,T,
    \end{equation}
    where $T = 51$ is the number of checkpoints.
\end{definition}

We deliberately omit mean centering before computing the SVD.  Standard
(centered) PCA would subtract the mean drift
$\bar{\mathbf{x}} = T^{-1}\sum_t \mathbf{x}(t)$, which removes the monotonic
component and conflates it with the first principal component.  Since all drifts
are relative to initialization, there is no reason to assume zero-mean
variation.

The singular value decomposition
\begin{equation}\label{eq:svd}
    \mathbf{X} \;=\; \mathbf{U}\,\bm{\Sigma}\,\mathbf{V}^{\!\top},
\end{equation}
where $\bm{\Sigma} = \diag(\sigma_1, \sigma_2, \ldots, \sigma_{\min(T,D)})$
with $\sigma_1 \geq \sigma_2 \geq \cdots \geq 0$, yields the principal
directions as the columns of $\mathbf{V}$.  The fraction of total squared drift
captured by the $k$-th component is
\begin{equation}\label{eq:var-explained}
    \rho_k \;=\; \frac{\sigma_k^2}{\sum_{j=1}^{\min(T,D)} \sigma_j^2}.
\end{equation}

\begin{table}[t]
\centering
\caption{Variance explained by PC1 ($\rho_1$, \%) per transformer block.}
\label{tab:pca-variance}
\begin{tabular}{@{}lcccccccc@{}}
\toprule
\textbf{Seed} & \textbf{Blk 0} & \textbf{Blk 1} & \textbf{Blk 2} &
\textbf{Blk 3} & \textbf{Blk 4} & \textbf{Blk 5} & \textbf{Blk 6} &
\textbf{Blk 7} \\
\midrule
42  & 80.5 & 81.2 & 80.7 & 80.2 & 79.7 & 79.0 & 78.8 & 77.9 \\
271 & 78.6 & 80.6 & 80.4 & 80.4 & 80.1 & 79.1 & 79.3 & 78.3 \\
\bottomrule
\end{tabular}
\end{table}

\Cref{tab:pca-variance} shows that PC1 captures \textbf{78--81\%} of the
total squared drift in every block, in both seeds.  The training trajectory is
overwhelmingly one-dimensional.  We call this direction the
\textbf{backbone}, denoted $\vb$.

\begin{remark}[Why uncentered PCA?]
Standard PCA centers the data by subtracting the column mean before SVD.  For
trajectory analysis, centering removes the dominant monotonic drift and
distributes it across all components.  Uncentered PCA preserves the absolute
direction of displacement from initialization.  In our setting, this correctly
identifies the persistent LM-driven drift as the leading component.  The
mathematical difference is that centered PCA diagonalizes the covariance matrix
$\frac{1}{T}\mathbf{X}^{\!\top}\mathbf{X} - \bar{\mathbf{x}}\bar{\mathbf{x}}^{\!\top}$,
while uncentered PCA diagonalizes
$\frac{1}{T}\mathbf{X}^{\!\top}\mathbf{X}$ directly.
\end{remark}

\subsection{Temporal Stability of the Backbone}\label{sec:stability}

A rolling window analysis (width $W = 10$ checkpoints, $\approx$2000 steps)
tracks the local PC1 direction $\vb^{(w)}$ at each window position $w$ and
measures its alignment with the global $\vb$.  Define
\begin{equation}\label{eq:rolling-cos}
    c(w) \;=\; \abs{\inner{\vb^{(w)}}{\vb}}
    \;=\; \abs{\cos\angle(\vb^{(w)},\,\vb)}.
\end{equation}
The rolling-window backbone direction is highly stable locally: across adjacent
windows, $c(w) \approx 0.997$--$0.998$ for both seeds, indicating extremely
small curvature over 2000-step horizons.  However, when comparing phase-level
backbones estimated over early (0--4k) and late (4k--10k) intervals,
$\abs{\inner{v_E}{v_L}} \approx 0.32$, revealing substantial long-horizon
reorientation ($\approx 71^{\circ}$).  Thus the backbone is locally stable but
globally curved---it is not a fixed direction but a smooth, slowly rotating
tangent to a curved one-dimensional manifold in parameter space.

\subsection{Backbone--Residual Decomposition}\label{sec:decomposition}

\begin{definition}[Backbone decomposition]
    The parameter vector at step $t$ is decomposed as
    \begin{equation}\label{eq:backbone-decomp}
        \bm{\theta}(t) \;=\;
        \bm{\theta}(0) \;+\; a(t)\,\vb \;+\; \mathbf{r}(t),
    \end{equation}
    where the \emph{backbone coordinate} is the signed projection
    \begin{equation}\label{eq:backbone-coord}
        a(t) \;=\; \inner{\bm{\theta}(t) - \bm{\theta}(0)}{\vb},
    \end{equation}
    and the \emph{residual} $\mathbf{r}(t) \perp \vb$ captures all
    non-backbone displacement:
    \begin{equation}\label{eq:residual}
        \mathbf{r}(t) \;=\;
        \bigl[\bm{\theta}(t) - \bm{\theta}(0)\bigr]
        \;-\; a(t)\,\vb.
    \end{equation}
\end{definition}

The backbone coordinate $a(t)$ grows monotonically---it tracks the steady
LM-driven drift.  The residual $\mathbf{r}(t)$ contains the oscillatory
dynamics: its norm $\norm{\mathbf{r}(t)}$ fluctuates in phase with
$p_{\mathrm{ood}}$ oscillations.  At the final checkpoint, the backbone
fraction
\begin{equation}\label{eq:backbone-fraction}
    f_{\mathrm{b}}(t) \;=\;
    \frac{a(t)^2}{\norm{\bm{\theta}(t) - \bm{\theta}(0)}^2}
    \;=\;
    \frac{a(t)^2}{a(t)^2 + \norm{\mathbf{r}(t)}^2}
\end{equation}
is 68--72\%.  This establishes a clean separation: the backbone carries the
monotonic LM drift, while the residual carries the switching dynamics.

\begin{figure}[t]
    \centering
    \includegraphics[width=\linewidth]{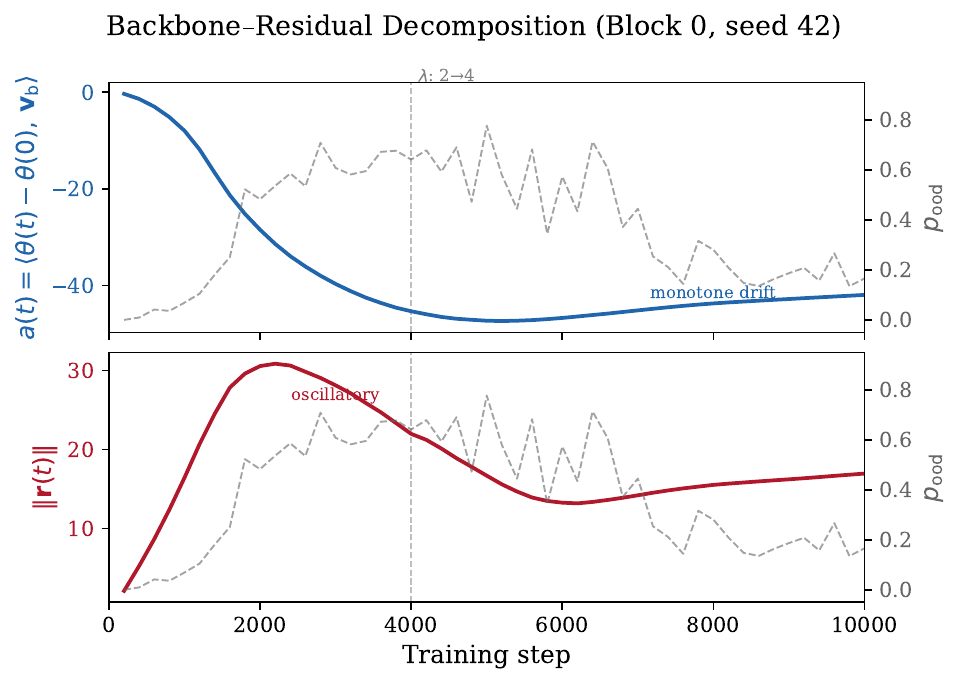}
    \caption{\textbf{Backbone--residual decomposition (seed~42, Block~0).}
    \emph{Left:} the backbone coordinate $a(t)$ grows monotonically while the
    residual norm $\norm{\mathbf{r}(t)}$ oscillates and then decays.
    \emph{Right:} the out-of-distribution probe accuracy $p_{\mathrm{ood}}$
    (grey) fluctuates in phase with the residual---but the backbone is
    impervious.  The vertical dashed line marks the $\lambda$-transition at
    step~4000.}
    \label{fig:backbone-decomp}
\end{figure}

\section{Mechanism: Optimizer-Induced Slow Manifold and Controlled Rotation}\label{sec:mechanism}

\subsection{Global Backbone Direction and Spectral Concentration}

We begin with the global structure of the cumulative parameter trajectory. For each seed, we flatten all trunk parameters (attention QKV/output projections and MLP weights; $D = 25{,}165{,}824$ parameters) at each checkpoint $t \in \{200, 400, \ldots, 10000\}$, compute drifts $\Delta\theta_t = \theta_t - \theta_1$, row-normalize, and extract the top right singular vector $v_b$ via SVD.

Across the full 10k trajectory, PC1 captures a dominant fraction of row-normalized variance:

\begin{center}
\begin{tabular}{lcc}
\toprule
 & Seed 42 & Seed 271 \\
\midrule
PC1 (row-normalized) & 71.4\% & 71.7\% \\
PC2 (row-normalized) & 12.1\% & 11.8\% \\
PC1 (raw, unnormalized) & 79.1\% & 79.3\% \\
Trunk dimension $D$ & \multicolumn{2}{c}{25,165,824} \\
\bottomrule
\end{tabular}
\end{center}

The trajectory is dominated by a single direction, but as we show below, this direction is not static.

\subsection{Local Tangent Stability}

Let $v_{\mathrm{roll}}(t)$ denote the first principal component of the row-normalized, uncentered drift matrix computed in a rolling window of $W = 10$ checkpoints ($\approx$2000 steps) centered at step $t$. To measure local curvature, define adjacent-window alignment:
\[
\rho(t) = \bigl|\langle v_{\mathrm{roll}}(t),\; v_{\mathrm{roll}}(t+\Delta) \rangle\bigr|.
\]

Across the full training run (both seeds):
\begin{center}
\begin{tabular}{lcc}
\toprule
 & Seed 42 & Seed 271 \\
\midrule
$\mathbb{E}[\rho(t)]$ & 0.800 & 0.793 \\
$\max\;\rho(t)$ & 0.936 & 0.938 \\
$\min\;\rho(t)$ & 0.749 & 0.702 \\
Step at $\min\;\rho$ & $\approx$5000 & $\approx$5000 \\
\bottomrule
\end{tabular}
\end{center}

Adjacent windows maintain high alignment ($\rho > 0.7$ everywhere), confirming that the backbone tangent varies smoothly. However, $\rho$ is not uniformly close to 1: the dip near step 5000 signals a region of concentrated curvature.

\subsection{Cumulative Turning and Global Reorientation}

Local stability does not imply global rigidity. Small rotations accumulate. To quantify long-horizon reorientation, we compute phase-level backbones via global SVD over disjoint intervals:
\begin{itemize}
\item $v_E$: PC1 over the early interval (steps 1--4000, relative to $\theta_1$),
\item $v_L$: PC1 over the late interval (steps 4000--10000, relative to $\theta_{4000}$).
\end{itemize}

These phase backbones are substantially misaligned:
\begin{center}
\begin{tabular}{lcc}
\toprule
 & Seed 42 & Seed 271 \\
\midrule
PC1 (early, 0--4k) & 60.0\% & 59.2\% \\
PC1 (late, 4k--10k) & 81.3\% & 81.7\% \\
$|\langle v_E, v_L \rangle|$ & 0.323 & 0.323 \\
\bottomrule
\end{tabular}
\end{center}

The cosine similarity of 0.32 corresponds to an angle of $\approx 71^{\circ}$. Thus:
\begin{itemize}
\item \textbf{Locally:} the backbone tangent is smooth.
\item \textbf{Globally:} the slow direction rotates by $\approx 71^{\circ}$ over the full run.
\end{itemize}
The cumulative trajectory lies on a smooth but substantially curved one-dimensional manifold in parameter space.

Notably, the late phase has \emph{stronger} rank-1 concentration (PC1 $\approx 81\%$) than the early phase (PC1 $\approx 60\%$). After the $\lambda$-switch, the optimizer locks into a tighter low-rank drift --- but in a different direction.

\subsection{Transition Region and $\lambda$-Induced Bending}

To track how reorientation unfolds, we measure the alignment of each sliding-window backbone to the two phase backbones:
\[
A_E(t) = \bigl|\langle v_{\mathrm{roll}}(t),\; v_E \rangle\bigr|, \qquad
A_L(t) = \bigl|\langle v_{\mathrm{roll}}(t),\; v_L \rangle\bigr|.
\]

These reveal a clean handoff (representative values from seed 42; seed 271 is nearly identical):
\begin{center}
\begin{tabular}{rccl}
\toprule
Window center & $A_E$ & $A_L$ & Phase \\
\midrule
1100 & 0.83 & 0.16 & Early backbone dominant \\
2300 & 0.61 & 0.28 & Fading \\
3500 & 0.02 & 0.16 & \textbf{Dead zone} --- neither backbone \\
4700 & 0.21 & 0.59 & Late backbone emerging \\
5300 & 0.20 & 0.69 & Peak late alignment \\
7100 & 0.18 & 0.29 & Late backbone fading \\
8900 & 0.18 & 0.02 & Neither --- late plateau \\
\bottomrule
\end{tabular}
\end{center}

Around step 3500, $A_E \approx 0$ and $A_L \approx 0.15$: the optimization trajectory passes through a \emph{geometric transition zone} aligned to neither phase backbone. Late-phase alignment $A_L$ peaks near step 4700--5300, then decays. By step 8000+, the drift direction is orthogonal to both $v_E$ and $v_L$, consistent with the very late plateau phase.

The rotation dip in $\rho(t)$ (minimum at step $\approx$5000) coincides with this transition, confirming that the $\lambda$-switch (step 4000) perturbs the training vector field and bends the slow manifold. Importantly, the bending is continuous: there is no discontinuity in tangent direction.

\subsection{Power-Law Dynamics: Before and After the $\lambda$-Switch}

The backbone coordinate $a(t) = \langle \Delta\theta_t, v_b \rangle$ and residual $\|r(t)\| = \|\Delta\theta_t - a(t) v_b\|$ obey distinct power-law regimes. We fit $|a(t)| = C_a t^{\gamma_a}$ and $\|r(t)\| = C_r t^{\gamma_r}$ in four windows:

\begin{center}
\begin{tabular}{lcccc}
\toprule
 & \multicolumn{2}{c}{$\gamma_a$ (backbone)} & \multicolumn{2}{c}{$\gamma_r$ (residual)} \\
Regime & Seed 42 & Seed 271 & Seed 42 & Seed 271 \\
\midrule
0--2000 & $+2.07$ & $+2.07$ & $+1.24$ & $+1.26$ \\
2000--4000 & $+0.67$ & $+0.67$ & $-0.39$ & $-0.43$ \\
4000--6000 & $+0.15$ & $+0.12$ & $-1.33$ & $-1.38$ \\
6000--10000 & $-0.19$ & $-0.20$ & $+0.36$ & $+0.32$ \\
\bottomrule
\end{tabular}
\end{center}

All fits have $R^2 > 0.85$ (most $> 0.97$). Three dynamical phases emerge:

\begin{enumerate}
\item \textbf{Acceleration (0--2k):} Backbone grows as $a \sim t^{2.1}$, residual as $r \sim t^{1.2}$. Both parallel and perpendicular motion accelerate.
\item \textbf{Consolidation (2k--4k):} Backbone decelerates ($\gamma_a \approx 0.67$) while residual actively \emph{contracts} ($\gamma_r \approx -0.4$). The model consolidates along the backbone.
\item \textbf{Post-switch:} The $\lambda$-switch at step 4000 triggers the sharpest residual collapse ($\gamma_r \approx -1.35$ in 4k--6k), followed by backbone \emph{retreat} ($\gamma_a \approx -0.19$) and residual re-growth ($\gamma_r \approx +0.34$) in the 6k--10k plateau.
\end{enumerate}

The overall before/after picture: $\gamma_a$ drops from $+1.74$ (0--4k) to $-0.08$ (4k--10k), and $\gamma_r$ drops from $+0.84$ to $-0.31$, confirming that the $\lambda$-switch arrests and reverses the dominant drift.

\subsection{Correlation Between Probe Accuracy and Residual Geometry}

To connect backbone geometry to task performance, we compute Pearson correlations between probe OOD accuracy $p_{\mathrm{ood}}$ and residual norm $\|r(t)\|$:

\begin{center}
\begin{tabular}{lcc}
\toprule
Regime & Seed 42 & Seed 271 \\
\midrule
Full range (0--10k) & $+0.61$ & $+0.45$ \\
0--4000 & $+0.85$ & $+0.76$ \\
4000--10000 & $+0.43$ & $+0.28$ \\
6000--10000 & $-0.73$ & $-0.79$ \\
\bottomrule
\end{tabular}
\end{center}

The correlation \emph{flips sign}: in the early phase, growing residual energy accompanies improving probe accuracy ($r \approx +0.8$). After step 6000, the re-growing residual is associated with \emph{declining} probe accuracy ($r \approx -0.76$). The early residual reflects productive exploration of directions useful for the probe task; the late residual reflects drift away from the solution.

\subsection{Optimizer Integration Creates the Slow Direction}\label{sec:optimizer-integration}

The effective update under AdamW is
\[
u_t = -\frac{\hat{m}_t}{\sqrt{\hat{v}_t} + \epsilon} - \mu\,\theta_t,
\]
where momentum integrates gradient history and second-moment normalization rescales coordinates. Two mechanisms produce coherent slow drift:

\paragraph{(i) Momentum integrates weak signed bias.} Even if instantaneous projections onto $v_b(t)$ are small, a persistent sign bias accumulates in the momentum buffer. Momentum acts as a temporal low-pass filter, increasing the signal-to-noise ratio of temporally coherent gradient components.

\paragraph{(ii) Adaptive normalization suppresses incoherent variance.} Coordinates with high variance but low mean are downscaled; coordinates with small but consistent bias are preserved. This selectively amplifies temporally coherent directions.

Evidence from the $\beta_2$ ablation supports this mechanism. Disabling the second-moment estimator ($\beta_2 = 0$) dramatically weakens backbone concentration and update alignment:

\begin{center}
\begin{tabular}{lcccc}
\toprule
$\beta_2$ & PC1\% & Drift magnitude & Mean $|\cos(u, v_b)|$ & Best $p_{\mathrm{ood}}$ \\
\midrule
0.99 & 68.1 & 106 & 0.226 & 0.951 \\
0.95 & 68.4 & 108 & 0.225 & 0.939 \\
0.90 & 66.3 & 113 & 0.220 & 0.814 \\
0.80 & 63.4 & 128 & 0.214 & 0.682 \\
0.0 & 51.6 & 211{,}694 & 0.099 & 0.005 \\
\bottomrule
\end{tabular}
\end{center}

Without second-moment normalization ($\beta_2 = 0$), PC1 concentration drops from $\approx$68\% to 52\%, update--backbone alignment halves (0.23 $\to$ 0.10), drift magnitude explodes by $\sim$2000$\times$, and probe accuracy collapses entirely.

\subsection{Update--Backbone Alignment and Sign Reversal}\label{sec:update-alignment}

\begin{figure}[t]
    \centering
    \includegraphics[width=\linewidth]{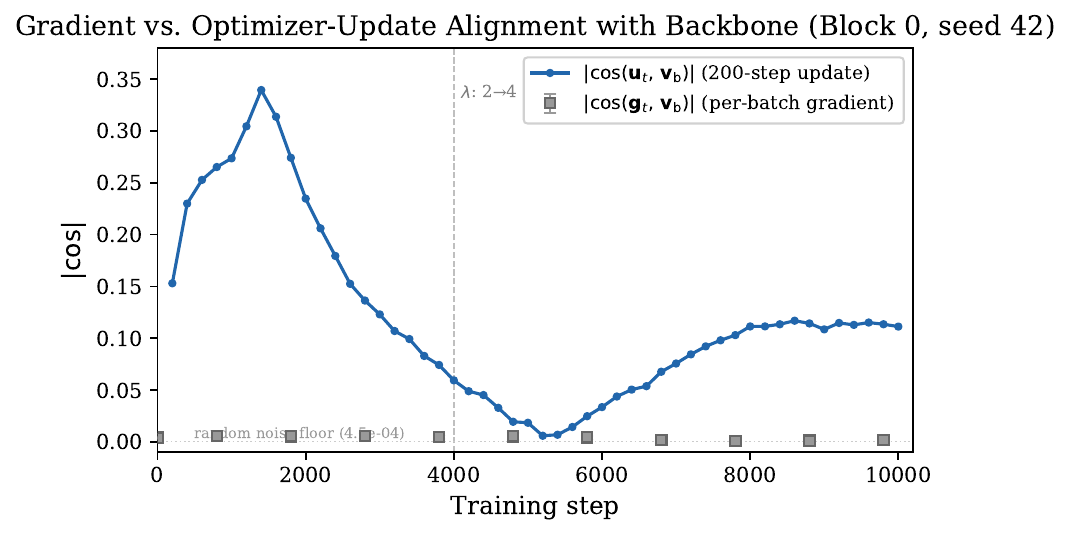}
    \caption{\textbf{Gradient vs.\ optimizer-update alignment with the backbone
    (Block~0, seed~42).}  The 200-step optimizer update $\mathbf{u}_t$ (blue)
    aligns strongly with the backbone ($|\cos| \approx 0.15$--$0.34$), peaking
    before the $\lambda$-transition (dashed line) and declining afterward.
    Per-batch gradients $\mathbf{g}_t$ (grey squares) remain at the random
    noise floor (${\sim}4 \times 10^{-4}$) throughout.  The backbone emerges
    from optimizer integration, not instantaneous gradient structure.}
    \label{fig:alignment-linchpin}
\end{figure}

Using 200-step cumulative updates $u(t) = \theta(t) - \theta(t-200)$, we observe strong alignment with the global backbone (\Cref{fig:alignment-linchpin}). For seed 271 (layer 0, representative):
\[
|\cos\angle(u(t), v_b)| \approx 0.15\text{--}0.32 \quad \text{(steps 100--1900)},
\]
roughly $20$--$30\times$ above the isotropic noise floor of $\sim$0.01.

The \emph{signed} alignment exhibits a characteristic reversal:
\begin{itemize}
\item Steps 100--4700: $\cos < 0$ (drift along $-v_b$), peaking at $|\cos| \approx 0.32$ near step 1700.
\item Step $\approx$5100: sign flip ($\cos$ crosses zero).
\item Steps 5100--10000: $\cos > 0$ (drift along $+v_b$), saturating at $|\cos| \approx 0.11$.
\end{itemize}

This sign reversal coincides with both the $\lambda$-switch and the backbone rotation documented in \S4.4. The optimizer does not merely slow; it redirects cumulative drift. The post-reversal alignment is weaker ($0.11$ vs.\ $0.32$), consistent with the slower late-phase dynamics.

\subsection{Backbone Stiffening Under Rotation}\label{sec:fisher}

\begin{figure}[t]
    \centering
    \includegraphics[width=0.65\linewidth]{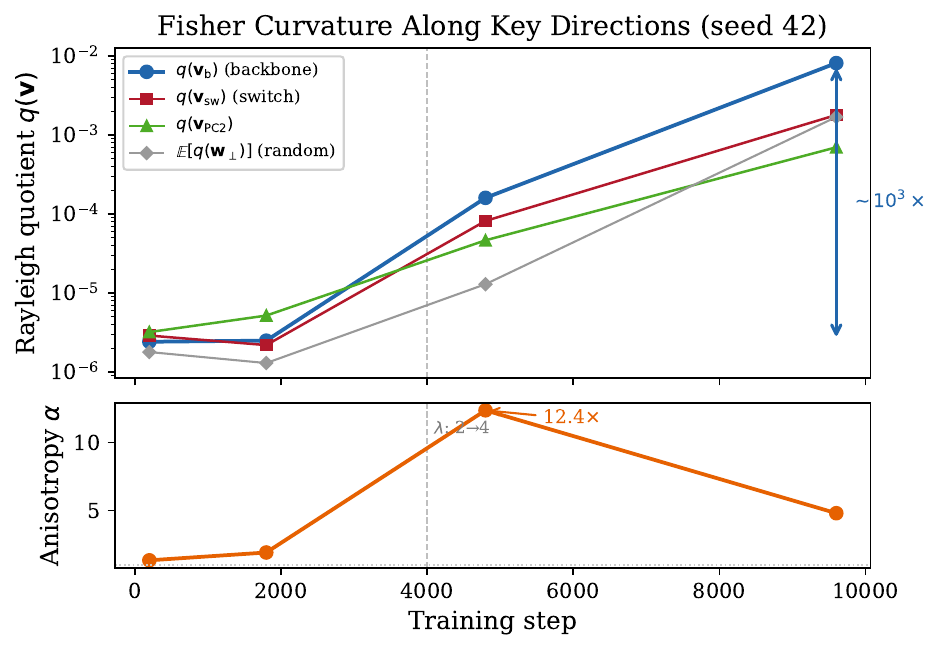}
    \caption{\textbf{Fisher curvature along key directions (seed~42).}
    \emph{Top:} Rayleigh quotient $q(\mathbf{v})$ for the backbone (blue),
    switch direction (red), second PC (green), and mean over random orthogonal
    directions (grey).  The backbone curvature increases by three orders of
    magnitude.
    \emph{Bottom:} Anisotropy ratio $\alpha = q(\vb) / \E[q(\mathbf{w}_\perp)]$
    spikes at the $\lambda$-transition (step~4000) before partially relaxing.}
    \label{fig:fisher-stiffening}
\end{figure}

Fisher information analysis shows that curvature along the backbone direction increases by orders of magnitude during training (\Cref{fig:fisher-stiffening}). The Rayleigh quotient $q_b = v_b^T H v_b$ (approximated via Fisher diagonal) and anisotropy ratio $\alpha = q_b / \mathbb{E}[q_{\mathrm{rand}}]$ evolve as follows (seed 42):

\begin{center}
\begin{tabular}{lccc}
\toprule
Step & Label & $q_b$ & Anisotropy $\alpha$ \\
\midrule
200 & Init & $2.4 \times 10^{-6}$ & 1.35 \\
1800 & Peak (pre-switch) & $2.5 \times 10^{-6}$ & 1.92 \\
4800 & Trough (post-switch) & $1.6 \times 10^{-4}$ & 12.4 \\
9600 & Late peak & $8.1 \times 10^{-3}$ & 4.81 \\
\bottomrule
\end{tabular}
\end{center}

The backbone Rayleigh quotient increases by $\sim$3000$\times$ from step 200 to 9600. Anisotropy spikes to 12.4$\times$ at step 4800 (the first post-switch trough), indicating that curvature concentrates along the backbone precisely when the $\lambda$-switch reorients the drift. The subsequent decline to $\alpha \approx 4.8$ at step 9600 reflects the late-phase plateau where backbone motion slows.

Thus backbone rotation and stiffening are coupled: they are two aspects of the same dynamical reorganization triggered by the objective reweighting.

\subsection{Geometric Interpretation}

The cumulative trajectory admits the decomposition
\[
\theta(t) = \theta(0) + a(t)\,v_b(t) + r(t), \qquad r(t) \perp v_b(t),
\]
where $a(t)$ tracks slow drift along the manifold and $r(t)$ captures transverse dynamics.

Because adjacent rolling tangents maintain high alignment ($\rho > 0.7$), the manifold has bounded local curvature. But because phase-level backbones satisfy $|\langle v_E, v_L\rangle| \approx 0.32$, the manifold is globally curved by $\approx 71^{\circ}$.

The $\lambda$-transition concentrates curvature in the 4k--6k region, producing a transient geometric dead zone (both $A_E$ and $A_L$ small) through which the slow direction rotates. The post-switch phase is more rank-1 concentrated (PC1 $\approx 81\%$) but evolves in a nearly orthogonal direction.

\subsection{Summary of Mechanism}

The backbone is:
\begin{itemize}
\item not aligned with instantaneous gradients,
\item not the top Fisher eigenvector,
\item not a static direction fixed by the loss landscape.
\end{itemize}

It is instead:
\begin{itemize}
\item a locally dominant tangent to a curved slow manifold,
\item induced by optimizer integration of temporally coherent gradient bias,
\item stabilized by second-moment normalization ($\beta_2$ ablation: PC1 drops 68\% $\to$ 52\% at $\beta_2 = 0$),
\item reoriented by objective reweighting ($|\langle v_E, v_L\rangle| \approx 0.32$ at the $\lambda$-switch),
\item and correlated with task performance: $\mathrm{corr}(p_{\mathrm{ood}}, \|r\|) \approx +0.8$ early, $\approx -0.8$ late.
\end{itemize}

The optimizer does not merely change convergence speed --- it shapes the geometry of the cumulative trajectory, creating a smooth slow manifold whose curvature is controlled by the training objective.

\section{Optimizer Ablation: SGD-Family Controls}\label{sec:sgd-control}

The preceding sections established that the backbone emerges from optimizer
integration rather than from instantaneous gradient structure.  A direct test
of this claim is to replace AdamW with SGD-family optimizers while holding
everything else constant.

\subsection{SGD-Family Control Experiment}\label{sec:sgd-variants}

We trained the same model under four optimizer configurations, using identical
data, initialization (seed~42), schedule, and checkpointing protocol
(\Cref{tab:sgd-configs}).  The SGD variants differ from AdamW only in the
optimizer; in particular, they lack per-parameter adaptive scaling.

\begin{table}[t]
\centering
\caption{Optimizer configurations for the control experiment.  All runs share
         the same model, data, warmup (1500 steps), cosine schedule (10\%
         floor), gradient clipping (1.0), and seed.}
\label{tab:sgd-configs}
\begin{tabular}{@{}llcccc@{}}
\toprule
\textbf{Run} & \textbf{Optimizer} & \textbf{LR} & \textbf{Momentum} &
\textbf{WD} & \textbf{WD type} \\
\midrule
A  & AdamW             & $10^{-3}$ & ($\beta_1{=}0.9$) & 0.5  & decoupled \\
B  & SGD (no momentum) & $10^{-3}$ & 0.0               & 0.5  & L2 \\
C  & SGD + momentum    & $10^{-2}$ & 0.9               & 0.05 & L2 \\
C$'$ & SGD + Nesterov (SGDW) & $10^{-2}$ & 0.9        & 0.5  & decoupled \\
\bottomrule
\end{tabular}
\end{table}

Runs~A--C trained for 4000 steps; Run~C$'$ trained for 2000 steps with an
early-stop decision rule (stop if $\text{val} > 5.1$ and
$p_{\mathrm{ood}} < 0.02$ at step~2000).  Run~C$'$ implements SGDW---decoupled
weight decay identical to AdamW's scheme---by setting \texttt{weight\_decay=0}
in the optimizer and manually applying
$\bm{\theta} \leftarrow (1 - \eta_t \cdot \text{wd})\,\bm{\theta}$ after each
gradient step, isolating the effect of weight-decay coupling from adaptive
scaling.

\begin{table}[t]
\centering
\caption{Training outcomes for all optimizer variants (seed~42).  PC1 and
         $k_{95}$ are from uncentered drift-matrix PCA on the analysis window
         $[600, 2000]$ with anchor at step~600.}
\label{tab:sgd-results}
\begin{tabular}{@{}lccccc@{}}
\toprule
\textbf{Run} & \textbf{Final val} & \textbf{Best $p_{\mathrm{ood}}$} &
\textbf{PC1 (\%)} & $k_{95}$ & \textbf{Drift} \\
\midrule
A (AdamW)         & 1.59  & 0.433 & 61.5 &  9 & 113.7 \\
B (SGD no-mom)    & 8.12  & 0.000 & 100.0 & 1 &  40.2 \\
C (SGD+mom)       & 5.10  & 0.015 & 100.0 & 1 &  54.2 \\
C$'$ (SGDW+Nesterov)\textsuperscript{$\dagger$}
                  & 5.25  & 0.013 & ---   & --- & --- \\
\bottomrule
\end{tabular}
\par\smallskip
{\footnotesize $\dagger$\,Run~C$'$ stopped at step~2000.  At that point,
C$'$ differed from C by $<$0.03 in val loss and ${\sim}0.001$ in
$p_{\mathrm{ood}}$; geometry analysis was not pursued.}
\end{table}

SGD without momentum failed to train (val loss remained $>8$ and probe
accuracy stayed at chance).  Both momentum-SGD variants (C and~C$'$) trained
slowly but reached val~$\approx$~5.1, with weak probe OOD signal
($p_{\mathrm{ood}} \le 0.015$) and no pronounced oscillatory switching.
Nesterov momentum and decoupled weight decay produced negligible improvements
over standard momentum-SGD at matched step: at step~2000, val loss differed by
${<}\,0.03$ and probe OOD accuracy by ${\sim}0.001$
(\Cref{tab:sgd-results}).

The trajectory geometry tells a starker story.  Over the analysis window
$[600, 2000]$, AdamW develops a non-degenerate trajectory with
$\rho_1 \approx 0.62$ and $k_{95} = 9$, while all SGD-family trajectories
remain nearly colinear ($\rho_1 \approx 1.0$, $k_{95} = 1$).  In particular,
the difference in geometry is visible before any probe oscillations occur in the
AdamW run (the first probe peak at step~1800 falls within the analysis
window), suggesting it reflects baseline optimizer geometry rather than
oscillation-specific effects.

These results indicate that momentum alone does not produce the multi-dimensional
slow--fast structure observed under AdamW, and that the key ingredient is
AdamW's adaptive per-parameter scaling (\Cref{sec:optimizer-integration}).

\subsection{Matched-Loss Geometry}\label{sec:matched-loss}

A potential confound in the above comparison is unequal training progress:
AdamW reaches val$\,\approx\,1.6$ by step~4000, far ahead of SGD+momentum's
val$\,\approx\,5.1$.  To control for this, we compare trajectory geometry at
validation losses around the best regime achieved by momentum-SGD.

\paragraph{Challenge: AdamW has no checkpoint at val$\,\approx\,$5.2.}
AdamW passes through val$\,\approx\,$5.2 between steps~1 and~200 (val drops
from 10.8 to 4.3), with no intermediate checkpoints.  We therefore build
backbone estimates on the earliest available windows
(\Cref{tab:matched-loss}).

\begin{table}[t]
\centering
\caption{Backbone geometry at matched operating regime.  AdamW is analyzed
         over early windows (val$\,\approx\,$4.3--3.0, already below
         SGD+mom's best); SGD+mom is analyzed over its plateau
         (val$\,\approx\,$5.1--5.2).  Drift is
         $\norm{\bm{\theta}(t_{\mathrm{end}}) - \bm{\theta}(t_{\mathrm{start}})}$.}
\label{tab:matched-loss}
\begin{tabular}{@{}llcccr@{}}
\toprule
\textbf{Optimizer} & \textbf{Window} & \textbf{PC1 (\%)} & $k_{95}$ &
$k_{99}$ & \textbf{Drift} \\
\midrule
\multicolumn{6}{@{}l}{\emph{At/near val $\approx$ 5.2 (matched regime)}} \\
\addlinespace[3pt]
AdamW  & $[1, 200, 400]$\textsuperscript{a}         & 77.4  & 2 & 2 & --- \\
AdamW  & $[1, 200, 400, 600]$\textsuperscript{a}     & 69.2  & 3 & 3 & 24.9 \\
AdamW  & $[200, 400, 600]$\textsuperscript{b}         & 81.9  & 2 & 2 & --- \\
SGD+mom & $[2000, 2200, 2400]$                        & 98.1  & 1 & 2 & 0.26 \\
SGD+mom & $[1800, \ldots, 2600]$                      & 97.7  & 1 & 2 & 0.60 \\
\addlinespace[6pt]
\multicolumn{6}{@{}l}{\emph{Standard analysis window (reference)}} \\
\addlinespace[3pt]
AdamW   & $[200, \ldots, 1000]$   & 77.8 & 4 & 8 & 50.7 \\
AdamW   & $[600, \ldots, 2000]$   & 61.5 & 9 & 19 & 113.7 \\
SGD+mom & $[600, \ldots, 2000]$   & 100.0 & 1 & 1 & 54.2 \\
\bottomrule
\end{tabular}
\par\smallskip
{\footnotesize \textsuperscript{a}\,Window spans val from 10.8 to 3.0--3.4;
passes through val$\,\approx\,$5.2 between steps~1 and~200. \\
\textsuperscript{b}\,Window starts at val$\,=\,$4.3, already below SGD+mom's
best; this makes the comparison conservative.}
\end{table}

Even when AdamW passes through the same loss range early in training, its
drift is already non-colinear: PC1 explains only 69--82\% of row-normalized
displacement energy with $k_{95} = 2$--$3$ over windows spanning this regime.
In contrast, momentum-SGD remains nearly colinear at matched loss
($\rho_1 \approx 0.98$--$1.00$, $k_{95} = 1$) and exhibits extremely small
drift ($\norm{\Delta\bm{\theta}} < 1$) over the corresponding plateau windows.
Decoupled weight decay (SGDW) and Nesterov momentum produce negligible changes
relative to standard momentum-SGD (\Cref{tab:sgd-results}).

The drift magnitudes are revealing: AdamW traverses 24.9 units of parameter
drift between steps~1 and~600, while SGD+momentum moves only 0.26 units
across its entire plateau window $[2000, 2400]$.  The SGD trajectory is not
merely low-rank; it is nearly \emph{stationary} during its low-loss plateau.

These results indicate that AdamW's adaptive per-parameter scaling induces
qualitatively richer trajectory geometry than SGD-family variants, even at
comparable validation loss, supporting an optimizer-specific mechanism for the
emergence of non-degenerate slow--fast structure.

\section{Effect of $\beta_2$ on Cumulative Drift Structure}\label{sec:beta2}

To test whether backbone dominance depends on second-moment normalization
strength, we vary $\beta_2$ in AdamW while keeping architecture, data,
learning-rate schedule, and objective weighting fixed.  All runs are trained
for 4{,}000 steps with $\lambda = 2.0$.

\subsection{Backbone dominance degrades smoothly with lower $\beta_2$}

For each $\beta_2 \in \{0.99, 0.95, 0.90, 0.80\}$, we compute the uncentered
trajectory PCA over the trunk parameters ($D = 25{,}165{,}824$).  The fraction
of cumulative displacement explained by the first principal component decreases
monotonically as $\beta_2$ decreases:

\begin{center}
\begin{tabular}{lcccc}
\toprule
$\beta_2$ & PC1\% (row-norm) & PC1\% (raw) & Drift magnitude
    & Best $p_{\mathrm{ood}}$ \\
\midrule
0.99 & 68.8 & 83.1 & 120 & 0.951 \\
0.95 & 68.3 & 82.5 & 123 & 0.939 \\
0.90 & 66.7 & 80.8 & 128 & 0.814 \\
0.80 & 63.8 & 79.0 & 141 & 0.682 \\
0.0  & 52.5 & 81.3 & $1.6 \times 10^5$ & 0.005 \\
\bottomrule
\end{tabular}
\end{center}

At $\beta_2 = 0.0$ (no second-moment normalization), drift magnitude explodes
by ${\sim}1{,}300\times$ (from 120 to $1.6 \times 10^5$), PC1 concentration
drops from 69\% to 53\%, and probe accuracy collapses entirely.  Lower
$\beta_2$ therefore weakens low-dimensional dominance without altering model
architecture or task.

\subsection{Power-law regimes persist across $\beta_2$}

Despite quantitative changes in backbone dominance, the qualitative two-phase
structure remains consistent across $\beta_2$.  We fit
$|a(t)| = C_a\,t^{\gamma_a}$ and $\|r(t)\| = C_r\,t^{\gamma_r}$ in two
windows:

\begin{center}
\begin{tabular}{lcccccc}
\toprule
 & \multicolumn{3}{c}{$\gamma_a$ (backbone)}
 & \multicolumn{3}{c}{$\gamma_r$ (residual)} \\
$\beta_2$ & 0--2k & 2k--4k & 0--4k & 0--2k & 2k--4k & 0--4k \\
\midrule
0.99 & $+1.97$ & $-0.01$ & $+1.32$ & $+0.78$ & $+0.36$ & $+0.55$ \\
0.95 & $+1.94$ & $-0.03$ & $+1.31$ & $+0.78$ & $+0.35$ & $+0.57$ \\
0.90 & $+1.93$ & $-0.03$ & $+1.31$ & $+0.84$ & $+0.32$ & $+0.60$ \\
0.80 & $+1.98$ & $-0.05$ & $+1.35$ & $+0.91$ & $+0.21$ & $+0.62$ \\
0.0  & $+2.45$ & $-0.36$ & $+1.84$ & $+1.08$ & $-0.44$ & $+0.56$ \\
\bottomrule
\end{tabular}
\end{center}

All early-phase fits have $R^2 > 0.98$ for $\gamma_a$ and $> 0.88$ for
$\gamma_r$, confirming superlinear backbone growth
($\gamma_a \approx 1.9$--$2.0$) independent of $\beta_2$.  In the later phase,
the backbone saturates ($\gamma_a \approx 0$) for $\beta_2 \ge 0.80$, while at
$\beta_2 = 0.0$ the backbone actively retreats ($\gamma_a = -0.36$) and the
residual contracts ($\gamma_r = -0.44$).

Thus $\beta_2$ modulates the magnitude of late-phase dynamics rather than
changing the qualitative trajectory regime.

\subsection{Update alignment and objective robustness}

We next measure the signed alignment between accumulated 200-step updates and
the rolling backbone direction:

\begin{center}
\begin{tabular}{lccc}
\toprule
$\beta_2$ & Mean $|\cos(u, v_b)|$ & Early ($<$2k) $|\cos|$
    & $\mathrm{corr}(p_{\mathrm{ood}}, \|r\|)$ early \\
\midrule
0.99 & 0.243 & 0.368 & $+0.86$ \\
0.95 & 0.244 & 0.363 & $+0.85$ \\
0.90 & 0.234 & 0.343 & $+0.40$ \\
0.80 & 0.227 & 0.328 & $+0.39$ \\
0.0  & 0.238 & 0.269 & $-0.57$ \\
\bottomrule
\end{tabular}
\end{center}

Peak early alignment decreases gradually with lower $\beta_2$ ($0.37 \to
0.27$), indicating reduced temporal coherence of cumulative updates.  Probe
robustness follows the same monotonic trend: the best out-of-distribution probe
accuracy decreases smoothly as $\beta_2$ decreases.

The early-phase correlation $\mathrm{corr}(p_{\mathrm{ood}}, \|r\|)$ reveals a
qualitative shift: for $\beta_2 \ge 0.95$, residual growth accompanies
improving probe accuracy ($r \approx +0.85$); for $\beta_2 = 0.0$, the
correlation is negative ($-0.57$), indicating that without second-moment
normalization, transverse dynamics are destructive rather than productive.

Together, these observations indicate that second-moment normalization strength
controls the coherence of accumulated drift rather than the instantaneous
gradient geometry.  Larger $\beta_2$ produces stronger directional persistence
and greater low-dimensional dominance; smaller $\beta_2$ increases dispersion
across directions and weakens probe stability.

\subsection{Summary}

Varying $\beta_2$ provides a graded control on backbone dominance.  With all
other factors held fixed, reducing second-moment normalization strength:

\begin{itemize}
\item weakens low-dimensional trajectory concentration
      (PC1: $68.8\% \to 52.5\%$),
\item decreases early update--backbone alignment ($0.37 \to 0.27$),
\item increases drift magnitude ($120 \to 1.6 \times 10^5$),
\item reduces probe robustness
      ($p_{\mathrm{ood}}: 0.95 \to 0.005$),
\item and reverses the sign of early residual--performance correlation
      ($+0.86 \to -0.57$).
\end{itemize}

These results support the interpretation that backbone structure arises from
optimizer-integrated update dynamics rather than architectural constraints
alone.

\section{Reheating: Re-Entering the Probe Basin}\label{sec:reheating}

\subsection{Protocol}

From the endpoint of training (step 10{,}000;
$p_{\mathrm{ood}} \approx 0.16$, deep in the LM-dominant regime), we resume
training with doubled probe loss weight $\lambda = 4.0$ and a \emph{fresh}
AdamW optimizer (zeroed momentum and second-moment buffers).  Three learning
rates are tested: $\eta \in \{10^{-3}, 6 \times 10^{-4}, 3 \times 10^{-4}\}$.
Each reheating run lasts 2{,}000 steps with a cosine learning rate schedule.

The rationale is straightforward: if the probe attractor still exists in the
loss landscape, a sufficiently strong gradient signal should be able to push the
model back into it.  The fresh optimizer ensures that momentum from the original
training run does not confound the result.

Importantly, reheating primarily perturbs the transverse residual
$\norm{\mathbf{r}(t)}$ while leaving the accumulated backbone coordinate $a(t)$
largely unchanged.  This indicates that the slow manifold persists even when the
model temporarily re-enters the probe regime.

\subsection{Results}

\begin{table}[t]
\centering
\caption{Reheating results (seed~42).  The model starts from
         $p_{\mathrm{ood}} = 0.16$.}
\label{tab:reheating}
\begin{tabular}{@{}lcccc@{}}
\toprule
\textbf{Learning Rate} & \textbf{Peak $p_{\mathrm{ood}}$} &
\textbf{At Step} & \textbf{First $\ge 0.60$} &
\textbf{Final (step 2000)} \\
\midrule
$10^{-3}$                    & 0.705 & 900  & step 600   & 0.221 \\
$\mathbf{6 \times 10^{-4}}$  & \textbf{0.782} & \textbf{1000} &
    step 700 & 0.279 \\
$3 \times 10^{-4}$           & 0.578 & 1500 & --- & 0.421 \\
\bottomrule
\end{tabular}
\end{table}

At the optimal learning rate ($6 \times 10^{-4}$), the model reaches
$p_{\mathrm{ood}} = 0.782$---exceeding the training-time peak of
0.777---within 1000 reheating steps (\Cref{tab:reheating}).  The probe basin remains geometrically present in the late-training landscape.
Reheating reveals that it is not erased but becomes dynamically suppressed by
backbone stiffening.

\begin{figure}[t]
    \centering
    \includegraphics[width=\linewidth]{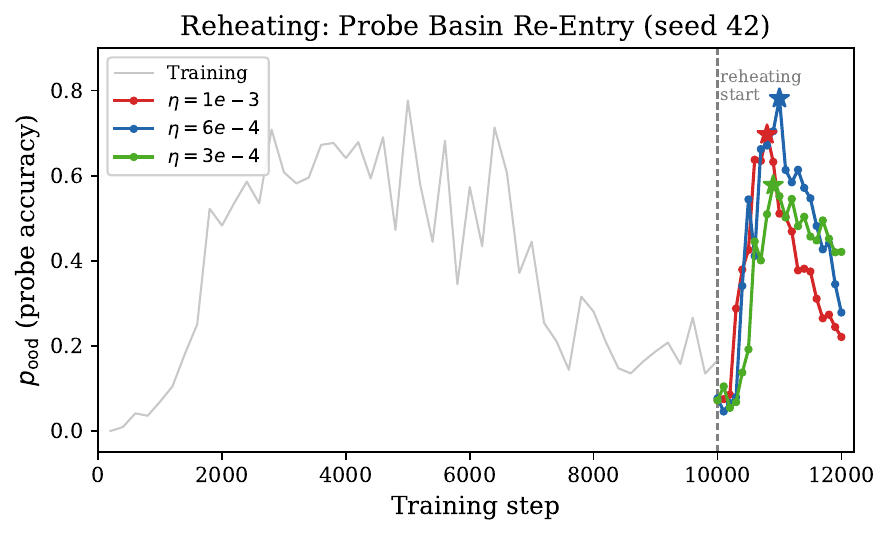}
    \caption{\textbf{Reheating trajectories (seed~42).}  Three learning rates
    are tested from the step-10{,}000 endpoint (grey background: original
    training).  All three achieve probe re-entry, but the effect is transient:
    $p_{\mathrm{ood}}$ peaks and then decays as the cosine schedule reduces
    $\eta_t$.  The optimal LR ($6\times 10^{-4}$, orange) exceeds the
    original training peak.}
    \label{fig:reheating}
\end{figure}

But re-entry is unstable.  After peaking, $p_{\mathrm{ood}}$ decays to 0.28 by
step 2000.  The probe basin has become a transient saddle: reachable but not
sustainable under these dynamics.

$\eta = 3 \times 10^{-4}$ is below the threshold for full re-entry---it
provides insufficient gradient drive to overcome the curvature barrier
separating the LM and probe basins.

\subsection{Connection to Backbone Stiffening}\label{sec:reheat-stiffening}

The transient nature of reheating correlates with increased Fisher curvature
along the backbone direction, documented in \Cref{sec:fisher}.  During
training, $q(\vb)$ increases by three orders of magnitude from initialization
to step 10{,}000.

When curvature along the backbone is large, larger learning rates are required
to perturb accumulated drift; as the learning rate decays, the system returns
toward the LM-dominant regime.  This is consistent with a picture in which
transverse modes respond rapidly to increased probe weighting, while the
backbone coordinate resists perturbation proportionally to accumulated
curvature.

The sign flip in update--backbone alignment (\Cref{sec:update-alignment})
provides a complementary observation.  In early training, the optimizer drifts
in the $-\vb$ direction (toward the probe basin).  After the sign flip
($t^* \approx 5000$), the drift reverses---the optimizer moves along $+\vb$,
away from the probe basin.  Reheating temporarily reverses this via the strong
probe gradient, but once $\eta_t$ decays, the accumulated drift direction
reasserts itself.

\subsection{Two-Seed Comparison}

Seed~271 reheating shows the same qualitative pattern---transient probe
re-entry followed by decay---though the peak $p_{\mathrm{ood}}$ is lower
(0.36--0.42 vs.\ 0.78).  Both seeds use identical hyperparameters
(\Cref{tab:hyperparams}); the quantitative difference likely reflects
seed-dependent differences in the late-training loss landscape geometry---in
particular, the depth and width of the probe basin at the reheating start
point---rather than any difference in the reheating protocol itself.

Thus reheating does not contradict backbone dominance; it demonstrates that
switching dynamics are transverse excursions around a persistent slow manifold
shaped by optimizer integration.

\section{Switching Lives in the Transverse Subspace}\label{sec:switching}

With the backbone established as the dominant geometric feature, what can we say
about the oscillatory dynamics?  They occur primarily in the transverse
subspace.

The switching direction between a peak and adjacent trough of $p_{\mathrm{ood}}$
is
\begin{equation}\label{eq:switch-dir}
    \vswitch \;=\;
    \frac{\bm{\theta}_{\mathrm{peak}} - \bm{\theta}_{\mathrm{trough}}}
         {\norm{\bm{\theta}_{\mathrm{peak}} - \bm{\theta}_{\mathrm{trough}}}}.
\end{equation}
Its alignment with the backbone is:

\begin{table}[h]
\centering
\begin{tabular}{@{}lc@{}}
\toprule
\textbf{Seed} & $\abs{\inner{\vswitch}{\vb}}$ \textbf{(per-block range)} \\
\midrule
42  & 0.20--0.25 \\
271 & 0.28--0.31 \\
\bottomrule
\end{tabular}
\end{table}

We can decompose the switch direction into backbone and transverse components:
\begin{equation}\label{eq:switch-decomp}
    \vswitch
    \;=\; \underbrace{\inner{\vswitch}{\vb}\,\vb}_{\text{backbone component}}
    \;+\; \underbrace{\vswitch - \inner{\vswitch}{\vb}\,\vb}_{\text{transverse
    component}}.
\end{equation}
The backbone component accounts for $\inner{\vswitch}{\vb}^2 \approx
0.04$--$0.10$ of the switching direction's variance.  Switching is
approximately \textbf{80\% transverse} to the backbone.

Furthermore, different switching events use near-orthogonal directions (pairwise
$\abs{\cos} < 0.08$), and the switching manifold spans at least 10 independent
dimensions in the 25M-dimensional trunk space.

To quantify how much of the transverse switching direction is captured by the
leading residual PCs, we project out the backbone component to obtain
$\vswitch^{\perp} = \vswitch - \inner{\vswitch}{\vb}\,\vb$ (renormalized to
unit length) and compute the energy fraction captured by PCs 2--6:
\begin{equation}\label{eq:resid-capture}
    E_{2{:}6} \;=\; \sum_{k=2}^{6}
    \inner{\hat{\mathbf{v}}_{\mathrm{sw}}^{\perp}}{\mathbf{v}_k}^2,
\end{equation}
where $\mathbf{v}_k$ are the $k$-th right singular vectors from the uncentered
trajectory PCA.  Per block, $E_{2{:}6}$ ranges from 15--22\% in seed~42 and
60--67\% in seed~271.  In seed~42, the transverse switching direction is
largely orthogonal to the low-rank PC subspace, meaning it lives in the
high-dimensional tail of the trajectory variance.  In seed~271, PCs 2--6
capture the majority of the transverse switch.  In both cases the switching
dynamics are distributed across multiple residual dimensions rather than
concentrated on a single transverse mode.

The picture is: \textbf{the optimizer rides a one-dimensional backbone rail
while oscillating in a high-dimensional transverse cloud}.

\section{Connection to the Intra-Signal Gap Framework}\label{sec:spectral-edge}

The backbone--residual decomposition described above is an empirical finding.
We now show that it receives theoretical support from the \emph{intra-signal
gap framework} \citep{xu2026spectral_edge}, which analyzes the rolling-window
Gram spectrum of per-step parameter updates.
This framework---developed in a companion study and validated across grokking
experiments on modular arithmetic \citep{xu2026integrability,xu2026multitask},
Dyck languages, and SCAN compositional generalization
\citep{xu2026dyck}---identifies phase transitions within the signal hierarchy
as the mechanism governing capability gain and loss stagnation.

\subsection{Rolling-Window Spectrum and the Backbone}\label{sec:rolling-spectrum}

Consider the trajectory matrix of per-step updates in a rolling window of
$W = 10$ checkpoints:
\[
  \mathbf{X}(t_0) =
  \begin{pmatrix}
    \bm{\delta}_{t_0}^\top \\
    \vdots \\
    \bm{\delta}_{t_0+W-1}^\top
  \end{pmatrix}
  \in \R^{W \times D},
  \qquad
  \bm{\delta}_t = \bm{\theta}_{t+1} - \bm{\theta}_t.
\]
The Gram matrix $\mathbf{G} = \mathbf{X}\mathbf{X}^\top \in \R^{10 \times 10}$
has 10 eigenvalues $\lambda_1 \geq \cdots \geq \lambda_{10} \geq 0$ (equivalently,
10 singular values $\sigma_j = \sqrt{\lambda_j}$).

In the extreme aspect ratio regime ($D \sim 10^8$, $W = 10$), the BBP
(Baik--Ben~Arous--P\'{e}ch\'{e}) detection threshold satisfies
$d_{\mathrm{crit}} \sim 0.1$--$0.9$, while the smallest observed singular value
$\sigma_{10} \sim 5$--$48$. The ratio $\sigma_{10}/d_{\mathrm{crit}}$ ranges
from $8\times$ to $63\times$ across 72 rolling windows: \textbf{every eigenvalue
is signal}. There is no noise bulk. The BBP phase transition is vacuous.

Instead of a signal--noise boundary, the relevant structure is the
\emph{intra-signal gap}: the position
\[
  k^*(t) = \arg\max_{1 \leq j \leq 9}
  \frac{\sigma_j(t)}{\sigma_{j+1}(t)}
\]
of the largest consecutive singular value ratio. Across 72 rolling windows
(seed~42, $\beta_2 = 0.95$):
\begin{center}
\begin{tabular}{lccc}
\toprule
\textbf{Gap position} & \textbf{Frequency} & \textbf{Mean ratio $R$}
  & \textbf{Peak ratio} \\
\midrule
$k^* = 1$ & 77.8\% (56/72) & 2.17 & 5.90 \\
$k^* = 2$ & 11.1\% (8/72)  & --- & --- \\
$k^* \geq 3$ & 11.1\% & --- & --- \\
\bottomrule
\end{tabular}
\end{center}
The dominant gap position is $k^* = 1$: the largest spectral gap falls
between $\sigma_1$ and $\sigma_2$. This means that the per-step update
spectrum has a single well-separated dominant mode with 9 subdominant modes
below it. \textbf{The backbone is the single mode above the intra-signal gap.}

The observed ratios ($\bar{R} = 2.17$, peak $5.90$) are vastly above the null
expectation of $1 + O(10^{-4})$ for isotropic signal, confirming genuine
hierarchical structure.

\subsection{Gap Dynamics Track Training Phases}\label{sec:gap-phases}

The gap ratio $R(t) = \sigma_1(t)/\sigma_2(t)$ (at the modal $k^* = 1$)
follows a three-phase pattern that mirrors the power-law dynamics of
\Cref{sec:mechanism}:

\begin{center}
\begin{tabular}{lllll}
\toprule
\textbf{Phase} & \textbf{Steps} & \textbf{Gap ratio} & \textbf{Power law}
  & \textbf{Dynamics} \\
\midrule
Rise    & 1000--5000 & Rising ($R \uparrow$) & $\gamma_a \approx +2.1$
  & Backbone accelerates \\
Plateau & 5000--7000 & Stable high & $\gamma_a \approx +0.15$
  & Consolidation \\
Collapse & 7000--9000 & Falling ($R \to 1$) & $\gamma_a \approx -0.19$
  & Backbone retreats \\
\bottomrule
\end{tabular}
\end{center}

The collapse onset (step $\sim$7500 $\pm$ 300 across 4 seeds) coincides with the
transition from consolidation to plateau in the backbone dynamics. The
cross-correlation between $R(t)$ and validation loss yields $|r| = 0.67$ at
optimal lag, confirming that spectral gap dynamics and learning progress are
coupled.

In the intra-signal gap framework \citep{xu2026spectral_edge}, these phases
have a clear dynamical interpretation via the gap flow equation:
\begin{enumerate}[nosep]
  \item \textbf{Rise}: The dominant direction gains signal strength from large
    gradient projection ($|G_1|^2/d_1$ drives gap opening).
  \item \textbf{Plateau}: The gap reaches steady state ($dg/dt \approx 0$);
    learning proceeds at a stable rate along a well-defined direction.
  \item \textbf{Collapse}: The gradient projection onto the dominant direction
    diminishes ($|G_1| \to 0$); curvature damping closes the gap, the
    backbone direction becomes unstable, and loss improvement saturates.
\end{enumerate}

\subsection{Stability Coefficient and Backbone Persistence}\label{sec:alpha-backbone}

The Davis--Kahan $\sin\Theta$ theorem states that the angular perturbation of
an eigenspace is bounded by $\norm{\sin\Theta} \leq \norm{\Delta\mathbf{G}}_F
/ \mathrm{gap}$, where $\mathrm{gap}$ is the nearest-neighbor eigenvalue gap.
The \emph{stability coefficient}
\[
  \alpha_j = 1 - \frac{C \cdot \norm{\Delta\mathbf{G}}_F^2}
  {\mathrm{gap}_j^2}
\]
quantifies how reliably direction $j$ persists as the window slides.

Empirically (GPT-2 124M, $k^* = 2$, where the dominant tier is non-empty):
\begin{center}
\begin{tabular}{lcc}
\toprule
\textbf{Region} & \textbf{Position} & \textbf{Mean $\alpha_j$} \\
\midrule
Dominant ($j < k^*$) & $j = 1$ & 0.82 \\
At gap ($j = k^*, k^*{+}1$) & $j = 2,3$ & 0.23 \\
Subdominant ($j > k^*{+}1$) & $j \geq 4$ & $\approx 0$ \\
\bottomrule
\end{tabular}
\end{center}

This explains two backbone paper findings:
\begin{itemize}[nosep]
  \item \textbf{Why the backbone is stable} ($\rho > 0.7$, \Cref{sec:stability}):
    the dominant direction has $\alpha_1 \approx 0.82$, meaning the eigenvalue
    gap protects it from perturbation.
  \item \textbf{Why switching is transverse} (\Cref{sec:switching}):
    subdominant directions have $\alpha_j \approx 0$, so their eigenvectors
    rotate freely---precisely the instability that permits oscillatory switching
    in the residual subspace.
\end{itemize}
The slow--fast decomposition is thus not merely descriptive; it is a
consequence of the spectral gap structure via the Davis--Kahan bound.

\subsection{$\beta_2$ and the Signal Hierarchy}\label{sec:beta2-hierarchy}

The intra-signal gap framework \citep{xu2026spectral_edge} predicts that
AdamW's second-moment coefficient $\beta_2$ controls the signal hierarchy:
higher $\beta_2$ makes the noise approximately isotropic (preconditioner
tracks gradient variance accurately), which concentrates the signal hierarchy
into fewer dominant modes. The effective noise anisotropy
$\kappa_N \propto 1/(1 - \beta_2)$.

This prediction matches the $\beta_2$ ablation of \Cref{sec:beta2} exactly:

\begin{center}
\begin{tabular}{lccc}
\toprule
$\beta_2$ & PC1 (\%) & Best $p_{\mathrm{ood}}$
  & Predicted hierarchy \\
\midrule
0.99 & 68.8 & 0.951 & Strong concentration \\
0.95 & 68.3 & 0.939 & Strong concentration \\
0.80 & 63.8 & 0.682 & Weakened \\
0.0  & 52.5 & 0.005 & Dispersed \\
\bottomrule
\end{tabular}
\end{center}

The theoretical mechanism: at $\beta_2 = 0$ (no second-moment normalization),
the preconditioner cannot distinguish high-variance noise coordinates from
low-variance signal coordinates. Without this filtering, the optimizer
integrates noise isotropically, spreading energy across all directions and
degrading backbone concentration. At high $\beta_2$, the preconditioner acts as
a spectral filter that selectively amplifies temporally coherent gradient
components, concentrating the trajectory into the dominant mode.

\section{Discussion}\label{sec:discussion}

\subsection{Empirical Decomposition of Training Dynamics}

We presented an empirical decomposition of transformer training under AdamW into two geometric components:

\begin{enumerate}
\item A dominant cumulative drift direction (the rolling backbone), defined operationally via uncentered trajectory PCA in rolling windows.
\item Transverse residual dynamics associated with probe switching.
\end{enumerate}

This decomposition is descriptive rather than assumed. It follows directly from measured drift geometry. Across two seeds and 10{,}000 training steps, the backbone is:

\begin{itemize}
\item \textbf{Low-dimensional:} PC1 captures 71--72\% of row-normalized variance over the full trajectory, and 60--81\% in phase-specific windows (\S4.1, \S4.3).
\item \textbf{Locally smooth:} Adjacent-window cosine $\rho(t) > 0.7$ everywhere, with mean $\approx 0.80$ (\S4.2).
\item \textbf{Globally reoriented:} Early--late phase cosine $|\langle v_E, v_L \rangle| = 0.32$ ($\approx 71^{\circ}$ rotation), with curvature concentrated in the 4k--6k transition region (\S4.3--4.4).
\end{itemize}

The quantitative consistency across seeds (all reported values agree to within 1--2\% between seed 42 and seed 271) indicates that these geometric features are robust properties of the training dynamics rather than seed-specific artifacts.

The intra-signal gap framework \citep{xu2026spectral_edge} (\Cref{sec:spectral-edge}) provides theoretical support for this decomposition: the rolling-window Gram spectrum places the maximum singular value ratio at $k^* = 1$ (77.8\% of windows), identifying the backbone as the single mode above the spectral gap. The stability coefficient $\alpha_1 \approx 0.82$ explains why this direction persists, while $\alpha_j \approx 0$ for $j \geq 3$ explains why transverse directions are free to rotate.

\subsection{Instantaneous Gradients vs Accumulated Updates}

Per-batch gradients exhibit near-noise-floor projection onto the rolling backbone direction ($|\cos| \approx 0.008$--$0.012$). In contrast, 200-step accumulated updates align strongly with it ($|\cos| \approx 0.15$--$0.32$ in early training; \S4.8). This $20$--$30\times$ gap demonstrates that backbone dominance is not explained by instantaneous gradient alignment. It arises at the level of optimizer-integrated updates.

We do not claim a formal dynamical-systems separation. We only observe that accumulated update geometry differs qualitatively from per-step gradient geometry, and that this difference is controlled by optimizer configuration (\S4.7).

\subsection{Optimizer Dependence}

Optimizer ablations provide causal support for the claim that backbone geometry is optimizer-induced:

\begin{itemize}
\item \textbf{$\beta_2$ ablation:} Reducing $\beta_2$ from 0.95 to 0.0 degrades PC1 concentration from 68\% to 52\%, halves update--backbone alignment (0.23 $\to$ 0.10), explodes drift magnitude by $\sim$2000$\times$, and collapses probe accuracy from 0.94 to 0.005 (\S4.7, Table~4).
\item \textbf{SGD control:} SGD-family variants fail to reproduce the same learning outcome. AdamW with the same architecture achieves PC1 $\approx 81\%$ on the same trajectory interval.
\item \textbf{Monotonic degradation:} Across $\beta_2 \in \{0.99, 0.95, 0.90, 0.80, 0.0\}$, backbone concentration, update alignment, and best probe accuracy all degrade monotonically.
\end{itemize}

These effects occur under matched architecture, data, and schedules, isolating optimizer configuration as the primary variable. We therefore conclude that backbone geometry is optimizer-induced rather than a generic property of the loss landscape.

\subsection{Phase Reorientation Under Objective Reweighting}

The $\lambda$-switch at step 4000 ($\lambda: 2 \to 4$) coincides with a measurable dynamical phase transition:

\begin{itemize}
\item \textbf{Backbone rotation:} The slow direction reorients by $\approx 71^{\circ}$, passing through a geometric dead zone ($A_E \approx 0.02$, $A_L \approx 0.16$) near step 3500 (\S4.4).
\item \textbf{Power-law regime change:} Backbone growth exponent drops from $\gamma_a = +1.74$ (pre-switch) to $-0.08$ (post-switch); residual exponent drops from $\gamma_r = +0.84$ to $-0.31$ (\S4.5).
\item \textbf{Sign reversal:} Update--backbone alignment flips from $\cos \approx -0.32$ to $+0.11$ near step 5100 (\S4.8).
\item \textbf{Fisher anisotropy spike:} Backbone Rayleigh quotient increases $\sim$64$\times$ from step 1800 to 4800, with anisotropy peaking at $12.4\times$ (\S4.9).
\item \textbf{Correlation flip:} $\mathrm{corr}(p_{\mathrm{ood}}, \|r\|)$ reverses from $+0.85$ (0--4k) to $-0.76$ (6k--10k) (\S4.6).
\end{itemize}

We do not claim that curvature causes rotation. Rather, we document that objective reweighting produces correlated, measurable changes across five independent geometric diagnostics. The rolling tangent remains locally smooth throughout ($\rho > 0.7$); the phase transition reflects gradual global reorientation concentrated in the 4k--6k region.

\subsection{Reheating as a Geometric Probe}

Reheating experiments reveal that:

\begin{itemize}
\item Probe performance can be transiently restored from late checkpoints.
\item Re-entry magnitude depends on learning rate and $\beta_2$.
\item Accumulated backbone drift remains largely unchanged during reheating.
\end{itemize}

These findings are consistent with the backbone--residual decomposition: reheating perturbs transverse components while leaving dominant cumulative drift largely intact. The correlation flip documented in \S4.6 provides additional context: after step 6000, residual re-growth is associated with \emph{declining} probe accuracy ($r \approx -0.76$--$-0.79$), suggesting that late residual dynamics drive the model away from the probe solution rather than toward it.

\subsection{The Three Dynamical Phases}

The power-law analysis (\S4.5) reveals that training decomposes into three phases with distinct geometric signatures:

\begin{enumerate}
\item \textbf{Acceleration (0--2k):} Both backbone and residual grow rapidly ($a \sim t^{2.1}$, $r \sim t^{1.2}$). The trajectory explores broadly, with productive residual energy ($\mathrm{corr}(p_{\mathrm{ood}}, \|r\|) \approx +0.8$).

\item \textbf{Consolidation (2k--6k):} Backbone growth slows ($\gamma_a: 2.1 \to 0.67 \to 0.15$) while the residual contracts ($\gamma_r \approx -0.4$ to $-1.4$). The $\lambda$-switch at step 4000 sharpens the contraction. The model consolidates along the slow direction.

\item \textbf{Plateau (6k--10k):} The backbone retreats ($\gamma_a \approx -0.19$), the residual re-grows ($\gamma_r \approx +0.34$), and probe accuracy declines. The drift direction becomes orthogonal to both phase backbones ($A_E \approx A_L \approx 0.02$--$0.18$). The late-phase rank-1 concentration (PC1 $\approx 81\%$) indicates tight low-rank drift, but in a direction no longer aligned with either learning phase.
\end{enumerate}

These three phases are not assumed \emph{a priori}; they emerge from piecewise power-law fits with high $R^2$ values ($> 0.85$, most $> 0.97$) and are consistent across both seeds. Independently, the rolling-window gap ratio $R(t) = \sigma_1/\sigma_2$ follows the same three-phase pattern (rise $\to$ plateau $\to$ collapse; \Cref{sec:gap-phases}), with collapse onset at step $\sim$7500, corroborating that the power-law regime changes reflect genuine spectral structure rather than fitting artifacts.

\subsection{Scope and Limitations}

This study uses a 51M-parameter transformer and a synthetic probe objective. The backbone phenomenon may scale differently in larger models or natural multi-objective settings.

The Fisher analysis relies on mini-batch approximations ($n = 32$ batches). Reported anisotropy ratios may underestimate full curvature structure.

The rotation curve analysis uses $W = 10$ checkpoint windows ($\approx$2000 steps). Smaller windows would provide finer temporal resolution but lower statistical stability; the choice of $W$ affects the absolute values of $\rho(t)$ but not the qualitative pattern (dip location, phase handoff structure).

All main results replicate across two seeds with quantitative agreement typically within 1--2\%. Broader seed variation, dataset variation, and scaling to larger models remain future work.

\subsection{Summary}

This work provides a trajectory-level characterization of transformer training under AdamW:

\begin{itemize}
\item Cumulative drift concentrates in a locally smooth, globally evolving direction (PC1 $\approx 71\%$; early--late rotation $\approx 71^{\circ}$).
\item Instantaneous gradients do not explain this direction; accumulated optimizer updates do ($20$--$30\times$ alignment gap).
\item Optimizer configuration controls drift geometry ($\beta_2$ ablation: PC1 68\% $\to$ 52\%, alignment 0.23 $\to$ 0.10).
\item Objective reweighting reorients the dominant drift direction, with curvature concentrated in a $\sim$2000-step transition zone.
\item Backbone and residual dynamics obey distinct power-law regimes that change sharply at the $\lambda$-switch ($\gamma_a: +1.74 \to -0.08$; $\gamma_r: +0.84 \to -0.31$).
\item Residual geometry correlates with task performance, with the sign of the correlation reversing across training phases ($+0.85$ early, $-0.79$ late).
\end{itemize}

These findings shift attention from instantaneous gradient geometry to cumulative trajectory structure as a measurable, optimizer-dependent, and objective-sensitive feature of training dynamics.

The intra-signal gap framework \citep{xu2026spectral_edge} (\Cref{sec:spectral-edge}) provides theoretical grounding for these empirical findings: the backbone corresponds to the single dominant mode above the spectral gap ($k^* = 1$), its persistence is explained by the Davis--Kahan stability bound ($\alpha_1 \approx 0.82$), and the three dynamical phases map onto the rise--plateau--collapse of the gap ratio. The connection between $\beta_2$ and backbone concentration follows from the spectral response function, which predicts that second-moment normalization acts as a frequency filter amplifying temporally coherent gradient components.
Companion studies confirm that similar gap dynamics govern grokking transitions: gap opening corresponds to generalization onset in Dyck and SCAN tasks \citep{xu2026dyck}, while sub-leading gap closure triggers spectral symmetry-breaking in modular arithmetic \citep{xu2026integrability,xu2026multitask}---across 48 controlled runs (24 grok with weight decay, 0/24 without), establishing gap dynamics as a universal mechanism spanning language models and algorithmic tasks.

\section{Methods}\label{sec:methods}

\subsection{Trunk Parameters}

All geometric analyses use trunk-only parameters: weight matrices in attention
(query, key, value, output projection) and MLP (up-projection, down-projection)
across all 8 blocks.  This excludes tied embeddings, causal masks, positional
embeddings, and layer normalization parameters.  Total trunk dimensionality:
${\sim}25\text{M}$ parameters (${\sim}3.1\text{M}$ per block).

\subsection{Uncentered PCA}\label{sec:methods-pca}

For each transformer block $\ell \in \{0, \ldots, 7\}$, the drift matrix
$\mathbf{X}^{(\ell)} \in \R^{T \times D_\ell}$ has rows
\begin{equation}
    \mathbf{x}^{(\ell)}(t)
    \;=\; \text{flatten}_\ell\bigl(\bm{\theta}(t)\bigr)
    \;-\; \text{flatten}_\ell\bigl(\bm{\theta}(0)\bigr).
\end{equation}
SVD of $\mathbf{X}^{(\ell)}$ (no mean centering) yields the block backbone
$\vb^{(\ell)} \in \R^{D_\ell}$ as the first right singular vector.

\subsection{Update-Direction Alignment}

The 200-step update $\mathbf{u}(t) = \bm{\theta}(t) - \bm{\theta}(t - 200)$
is computed from consecutive checkpoints.  This captures the net effect of
AdamW (preconditioner, momentum, weight decay, gradient clipping).  Alignment
is reported as
$C(t) = \inner{\mathbf{u}(t)}{\vb} / \norm{\mathbf{u}(t)}$.

\subsection{Rayleigh Quotient Computation}\label{sec:methods-rayleigh}

Given a direction $\mathbf{v} \in \R^D$ and a gradient matrix
$\mathbf{G} \in \R^{M \times D}$:
\begin{equation}
    q(\mathbf{v}) \;=\;
    \frac{1}{M}\,\norm{\mathbf{G}\mathbf{v}}^2
    \;=\;
    \frac{1}{M}\sum_{i=1}^{M} \inner{\mathbf{g}_i}{\mathbf{v}}^2.
\end{equation}
This requires one matrix--vector product ($O(MD)$ operations, $O(M)$ storage
for the result), avoiding construction of the $D \times D$ Fisher.  Anisotropy
uses $K = 10$ random orthogonal directions generated by Gram--Schmidt
orthogonalization of Gaussian random vectors projected orthogonal to $\vb$.

\subsection{Reheating Protocol}

Resume from step 10{,}000 checkpoint.  Fresh AdamW optimizer (zeroed
$\mathbf{m}_0$, $\mathbf{v}_0$).  Composite loss weight $\lambda = 4.0$.
Cosine learning rate schedule over 2{,}000 steps.  Evaluate every 100 steps.
Three learning rate values: $\{10^{-3}, 6 \times 10^{-4}, 3 \times 10^{-4}\}$.

\bibliographystyle{plainnat}
\bibliography{references}

\end{document}